\documentclass[10pt]{cai}

\usepackage{algorithm, algorithmic}
\newcommand{\CL}[1]{
	\STATE \textcolor{blue}{// \textit{#1}}
}
\usepackage{listings}
\usepackage{array}
\usepackage{multirow}
\newcolumntype{P}[1]{>{\centering\arraybackslash}p{#1}}
\newcolumntype{M}[1]{>{\centering\arraybackslash}m{#1}}
\newcommand*{\percenting}[0]{\%}
\usepackage{makecell}

\usepackage{siunitx}
\newcommand\rdnn {\iftrue\expandafter\rdd\expandafter{\else}\fi}
\newcommand\rdtt {\iftrue\expandafter\rdt\expandafter{\else}\fi}
\newcommand\rdtb {\iftrue\expandafter\rddbft\expandafter{\else}\fi}
\newcommand\rdbf {\iftrue\expandafter\rddbf\expandafter{\else}\fi}
\newcommand{\f}{4}
\newcommand{\ft}{2}
\newcommand\rdd [1]{ \num[round-mode=places,round-precision=\f]{#1}\egroup}
\newcommand\rdt [1]{ \num[round-mode=places,round-precision=\ft]{#1}\egroup}
\newcommand\rddbf [1]{\bfseries \num[round-mode=places,round-precision=\f]{#1} \egroup}
\newcommand\rddbft [1]{\bfseries \num[round-mode=places,round-precision=\ft]{#1} \egroup}

\usepackage[only,varodot]{stmaryrd}
\SetSymbolFont{stmry}{bold}{U}{stmry}{m}{n}

\usepackage{amsfonts}
\usepackage{mathtools}

\usepackage{float}

\usepackage{csquotes}
\usepackage{amsmath}

\usepackage[most]{tcolorbox}
\newtcolorbox[auto counter]{remarkbox}[2][]{lower separated=true,
	drop fuzzy shadow,
	boxrule=0.6pt,
	colback=white,
	colframe=black,
	coltitle=black,
	enhanced,
	boxed title style={colframe=white,colback=white,left=0pt,right=0pt},
	attach boxed title to top left={xshift=0.25cm,yshift=-2.5mm},
	title=\scriptsize\textsc{\textbf{Remark \thetcbcounter}} #2, #1}

\newtcolorbox[auto counter]{propbox}[2][]{lower separated=true,
	drop fuzzy shadow,
	boxrule=0.6pt,
	colback=white,
	colframe=black,
	coltitle=black,
	enhanced,
	boxed title style={colframe=white,colback=white,left=0pt,right=0pt},
	attach boxed title to top left={xshift=0.25cm,yshift=-2.5mm},
	title=\scriptsize\textsc{\textbf{Proposition \thetcbcounter}} #2, #1}

\newtcolorbox[auto counter]{assumptionbox}[2][]{lower separated=true,
	drop fuzzy shadow,
	boxrule=0.5pt,
	colback=white,
	colframe=black,
	coltitle=black,
	enhanced,
	boxed title style={colframe=white,colback=white,left=0pt,right=0pt},
	attach boxed title to top left={xshift=0.25cm,yshift=-2.5mm},
	title=\scriptsize\textsc{\textbf{Assumption \thetcbcounter}} #2, #1}

\newcommand{\textmc}[1]{\texttt{#1}}

\newcommand{\textccpt}[1]{\texttt{#1}}

\begin{document}
\def\conferenceyear{2024}
\volumeheader{37}{0}
\begin{center}

	\title{Guided AbsoluteGrad : Magnitude of Gradients \\Matters to Explanation's Localization and Saliency}
	\maketitle

	\thispagestyle{empty}

	\begin{tabular}{cc}
		Jun Huang\upstairs{\affilone,*}, Yan Liu\upstairs{\affilone}
		\\[0.25ex]
		{\small \upstairs{\affilone} Concordia University, Montreal, Canada} \\
	\end{tabular}

	\emails{
		\upstairs{*}jun.huang@mail.concordia.ca
	}
	\vspace*{0.2in}
\end{center}

\begin{abstract}
	This paper proposes a new gradient-based XAI method called \textit{Guided AbsoluteGrad} for saliency map explanations.
	We utilize both positive and negative gradient magnitudes and employ gradient variance to distinguish the important areas for noise deduction.
	We also introduce a novel evaluation metric named \textmc{ReCover And Predict (RCAP)}, which considers the \textccpt{Localization} and \textccpt{Visual Noise Level} objectives of the explanations.
	We propose two propositions for these two objectives and prove the necessity of evaluating them.
	We evaluate \textit{Guided AbsoluteGrad} with seven gradient-based XAI methods using the \textccpt{RCAP} metric and other SOTA metrics in three case studies:
	(1) ImageNet dataset with ResNet50 model;
	(2) International Skin Imaging Collaboration (ISIC) dataset with EfficientNet model;
	(3) the Places365 dataset with DenseNet161 model.
	Our method surpasses other gradient-based approaches, showcasing the quality of enhanced saliency map explanations through gradient magnitude.
\end{abstract}

\copyrightnotice

\pagestyle{empty}

\section{Introduction}
\label{sec:intro}
Saliency map~\cite{vanillagrad} is commonly used to explain image classification tasks in computer vision.
It highlights the pixels according to their contribution to the prediction.
The work~\cite{methodcate} divides saliency methods into three branches: CAM-based (Class Activation Map), gradient-based, and perturbation-based methods.
CAM-based methods such as Grad-CAM~\cite{gradcam} and GradCAM++~\cite{gradcampp} are commonly adopted, relying on access to the model's feature map and gradient accumulation.
These methods produce smoothed saliency maps of low resolution from the feature map.
Gradient-based methods such as Vanilla Gradient~\cite{vanillagrad} and SmoothGrad~\cite{smoothgrad}, on the other hand, leverage only the gradient values of the input as the explanations.
These methods reduce explanation noise by introducing randomness to the sample multiple times and aggregating the gradients into one saliency map.
Other gradient-based methods such as Integrated Gradients~\cite{ig}, Guided Integrated Gradients~\cite{guidedig}, and BlurIG~\cite{blurig} generate saliency maps by building a path between the target and baseline images.
The Guided Backpropagation~\cite{guidedbp} calculates the saliency map by modifying the model's backpropagation process to exclude the negative gradients on ReLU activation.
Perturbation-based methods such as RISE~\cite{RISE} and SHAP~\cite{shap} calculate feature contribution by analyzing the prediction output and multiple modified inputs.
In this paper, we focus on developing gradient-based XAI methods for neural network models.

Interpreting the negative gradient remains a challenge during saliency map generation.
The works in~\cite{guidedbp, smoothgrad, gradcam, ig} assume that the negative gradient is either noise that causes countereffect or its effect depends on the contexts of datasets and models.
The practices usually exclude negative gradients during the backpropagation or sum them with the positive gradients when generating the aggregated saliency map.
In contrast, the study in~\cite{bigrad} indicates that negative gradients also play a significant role in feature attribution on remote sensing imaging.
The works in~\cite{xai_servey_2, roar} make a similar hypothesis on negative gradients.
The above works lack comprehensive discussions and experiments on how negative gradients affect the saliency map on difference gradient-based methods.
We aim to identify the position of the negative gradient and further leverage it to generate better saliency map explanations.
Our contribution to this paper can be summarized in the following threefold.

\textbf{Firstly}, we propose a gradient-based method called \textit{Guided AbsoluteGrad} that considers the magnitude of positive and negative gradients.
Our method works as follows:
(1) for each sample, we obtain multiple saliency maps from different modifications of the original image;
(2) we absolutize the gradient values to measure the gradient magnitude as the feature attribution;
(3) we calculate the gradient variance as a guide to the target area;
(4) we sum the absolute gradients multiplied by the guide as the explanation.

\textbf{Secondly}, we identify the limitation of the existing saliency map evaluation techniques and define the \textmc{ReCover And Predict (RCAP)} that focuses on \textccpt{Localization} and \textccpt{Visual Noise Level} objectives of the explanations.
We collect multiple recovered images on a baseline image based on different partitions of the ranked saliency map values.
We get the prediction scores of these recovered images.
For each partition and its recovered image, we multiply
(1) the ratio saliency value of the partition against the entire explanation,
and (2) the prediction score of the recovered image as the evaluation result.

\textbf{Finally}, we conduct experiments involving three datasets, three models, and seven state-of-the-art (SOTA) gradient-based XAI methods.
The result shows that, for \textccpt{RCAP} and other SOTA metrics, the Guided AbsoluteGrad method outperforms other gradient-based XAI methods.
We also propose two propositions and provide experiments to discuss and validate how \textccpt{RCAP} evaluates the two objectives.
The code of this paper is released in GitHub\footnote{\url{https://github.com/youyinnn/Guided-AbsoluteGrad}}.

\section{Related Works}
\label{sec:rw}
In this section, we discuss the existing negative gradient interpretations and our assumptions based on them.
We also discuss the saliency map evaluation metrics and their drawbacks, therefore proposing a new metric.

\subsection{Interpretation of Negative Gradients.}
\label{sec:rw_nega_grad}
We summarize three groups of negative gradient interpretations from the existing works.

\textit{Group 1: Negative Gradient Can Not Indicate Target Class Attribution.}
Guided Backpropagation~\cite{guidedbp} argues that the negative gradients stem from the neurons deactivated regarding the targeted class;
Grad-CAM~\cite{gradcam} thinks that negative gradients might belong to other categories.
Both methods rule out the negative gradient from their saliency explanations.
Based on~\cite{ig}, the XRAI~\cite{xrai} states that regions unrelated to the prediction should have near-zero attribution,
and regions with competing classes should have negative attribution.
That is saying the negative gradient is not used to explain the target class.

\textit{Group 2: The Indication of Negative Gradient Depends on the Data Context.}
SmoothGrad~\cite{smoothgrad} suggests that whether to keep the negative gradient depends on the context of the dataset or individual sample.
DeepLift~\cite{deeplift} addresses the limitations of~\cite{vanillagrad, guidedbp, ig} by considering the effects of positive and negative contributions at nonlinearities separately.

\textit{Group 3: Negative Gradient Indicates Target Class Attribution.}
The work in ~\cite{xai_servey_2} hypothesizes that negative gradients form around the important features and further complement the positive gradient for better visualization.
The study~\cite{roar} found that the magnitude may be far more telling feature importance than the direction of the gradient.
Another work~\cite{bigrad} provides supporting experiments and proposes the BiGradV, which verifies that the negative gradients also play a role in saliency map explanation in the remote sensing images.

The above interpretations are limited to the direction of the gradient based on their own XAI methods, 
handling negative gradients differently.
There is a lack of empirical experiments on how the negative gradients solely affect the explanations.
With this regard, we propose an assumption that could be applied to any gradient-based method, that is, \textit{the magnitude of both positive and negative gradients matters to feature attribution}.
Based on the assumption, we develop a gradient-based XAI method that leverages the gradient magnitude.
Moreover, we modify the existing gradient-based methods by taking the negative gradient direction and conducting empirical experiments to verify our assumption.

\subsection{Existing Saliency Map Evaluation Metric}
\label{sec:rw_eval}
The \textmc{Log-Cosh Dice Loss}~\cite{logcoshdiceloss} evaluates the loss between the ground-truth segmentation and the saliency map regarding the target object.
Similarly, the work in~\cite{blurig} uses \textmc{Mean Absolute Error (MAE)} to evaluate the performance based on ground-truth segmentation.
The works of~\cite{smoothgrad, gradcam} employ humans to evaluate their work with questionnaires.
However, collecting ground-truth segmentation or conducting human studies, in most cases, is unavailable and expensive.
\textmc{Deletion and Insertion Area Under Curve (DAUC\&IAUC)} are widely adopted~\cite{RISE,xrai,guidedig,groupcam} to evaluate the saliency map by getting the prediction from the gradually removing or inserting the important pixels of the input.
Likewise, the \textmc{RemOve And Retrain (ROAR)}~\cite{roar} evaluates the saliency map by assessing the performance of the retrained model on the modified datasets.
Such modified datasets are generated by gradually removing features based on saliency.
However, the \textmc{ROAR} lacks the theoretical and empirical basis to justify the necessity of retraining.
We argue that the retraining process brings more uncertainty due to different training strategies that involve hyperparameter optimization, machine learning model selection, and data processing strategies.
Existing work~\cite{roar_crit} also reveals the contradictory evaluation result against \textmc{ROAR} and discusses that the mutual information between the input and the target label can lead to biased attribution results during the \textmc{ROAR} retraining.
Moreover, both \textmc{DAUC\&IAUC} and \textmc{ROAR} evaluate how accurately the explanation can locate the important features but ignore the visual noise level of the explanations.
Therefore, we propose \textmc{ReCover And Predict (RCAP)} that focuses on the \textccpt{Localization} and \textccpt{Visual Noise Level} objectives at the same time.
We define propositions followed by theoretical proof and conduct experiments to describe 
how \textmc{RCAP} guarantees these objectives are evaluated correctly.

\section{Define Guided AbsoluteGrad}
\label{sec:method}
In this section, we summarize the algorithm from the SOTA gradient-based methods into three parts and present the existing implementation of each part.
Then, we present how we leverage the gradient magnitude to define the proposed method.

\subsection{Gradient-based XAI Method Abstraction}
\label{sec:method_proposed}
In general, the gradient-based methods are based on the Vanilla Gradient~\cite{vanillagrad} and can be summarized into three parts:
(1) Input Modification, (2) Gradient Interpretation, and (3)  Gradient Aggregation from multiple modifications.
We represent the input modification with the generated saliency map as follows:
\begin{equation}
	\label{eq:grad}
	M_c(x, \gamma) = \nabla f_c(\gamma(x)) = \frac{\partial f_c(\gamma(x))}{\partial (\gamma(x))}
\end{equation}
where $f_c(x)$ is the activation function regarding class \textit{c} of the input image \textit{x} and $\gamma$ is the modification function for saliency explanation noise removal.
SmoothGrad~\cite{smoothgrad} achieves this by introducing the Gaussian noise to the input.
Integrated Gradients~\cite{ig} samples multiple modifications from the evolution path between the baseline image and the original input.

For gradient interpretations, we can rule out the negative gradient~\cite{guidedbp,gradcam} or take the summation~\cite{smoothgrad}, considering the negative gradient as the disturbance of explanations.
We can also take the absolute gradient, leveraging the magnitude as feature importance.

And finally, there are different aggregation strategies to generate the final saliency map explanations.
Suppose $M_c^n(x, \gamma) = \{M_c(x, \gamma_i)\ |\ i \in 1,2,\dots,n\}$ indicates the resulting  $n$ variant of the saliency maps and the $\gamma_i$ is the $i^{th}$ modification of function $\gamma$.
We can take the saliency mean~\cite{smoothgrad}, considering the average saliency represents the best of the explanation.
We can also use the saliency variance~\cite{vargrad}, leveraging the magnitude of saliency values' changes as explanations.

\subsection{Leveraging the Gradient Magnitude by Absolutization and Variance Guide}
\label{sec:gag_def}
Now, we define our Guided AbsoluteGrad method.
We first define AbsoluteGrad by absolutizing the gradient values and taking the mean aggregation:
\begin{equation}
	\label{eq:proposed_0}
	M^{\textrm{AG}} = \overline{|M_c^n|}(x, \gamma) = \frac{1}{n} \displaystyle\sum_{i=1}^n |M_c(x, \gamma_i)|
\end{equation}

Since the absolutization might also magnify the saliency value of the features that do not represent the target object,
we leverage the variance of $M_c^n(x, \gamma)$ to filter the features.
This is based on the idea that features with fierce gradient variation determine if they are important or not.
The more important the feature is, the larger the gradient variation from negative to positive due to the modification.
We calculate the gradient variance with:
\begin{equation}
	\label{eq:variance}
	\mathbb{V}(M_c^n(x, \gamma)) = \frac{1}{n - 1} \sum_{i=1}^n (M_c(x, \gamma_i) - \overline{M_c^n}(x, \gamma))^2
\end{equation}

We define a \textbf{variance guide} $\mathbb{G}$ to filter the feature based on a threshold percentile $p$.
\begin{equation}
	\label{eq:norm_variance_processing}
	\mathbb{G} = \begin{cases}
		1                            & \mathbb{V}(M_c^n(x, \gamma)) \ \ge\ p^{th} \ \text{percentile} \\
		\mathbb{V}(M_c^n(x, \gamma)) & \text{otherwise}
	\end{cases}
\end{equation}

Finally, we multiply the variance guide $\mathbb{G}$ with $M^{\text{AG}}$ as the saliency explanations.
In summary, the absolute gradient and the mean aggregation determine how much contribution the feature has; the gradient variance guide determines if the feature is important or not.

\begin{figure}[ht]
	\centering
	\includegraphics[width=\linewidth]{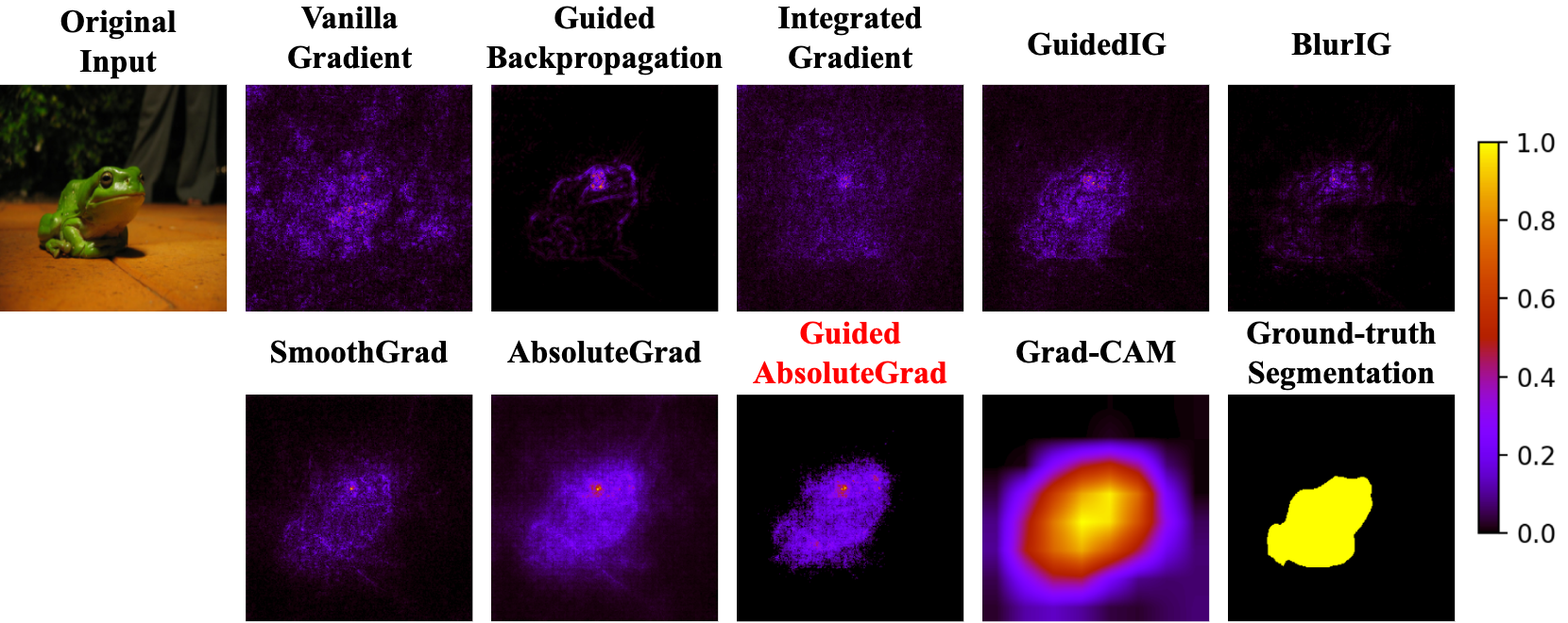}
	\caption{
		Illustrations of the saliency map explanations of the Guided AbsoluteGrad method and other SOTA methods.
		The saliency value is normalized, and the color bar on the right indicates the mapping of colors and values.
	}
	\label{fig:xai_example3}
\end{figure}

Algorithm \ref{alg:x} and Figure~\ref{fig:gag_process} describe the detailed implementation of the above process.
The algorithm can leverage any modification function.
We conduct experiments on XAI methods with different modifications:
SmoothGrad~\cite{smoothgrad}, Integrated Gradients~\cite{ig}, Guided Integrated Gradients~\cite{guidedig}, and BlurIG~\cite{blurig}.
We find that introducing the Gaussian noise performs the best for stimulating the gradient variation.
Hence, in this paper, we add the Gaussian noise to modify the input.
Figure \ref{fig:xai_example3} shows the examples of saliency explanations from the Guided AbsoluteGrad method with $p = 85^{th}$ and other SOTA gradient-based methods.
We also include the CAM-based method of Grad-CAM~\cite{gradcam}.
Our method explains high saliency to the area of the frog and low saliency to the background.

\begin{minipage}[t]{0.43\linewidth}
	\begin{algorithm}[H]
		\caption{Guided AbsoluteGrad}
		\label{alg:x}
		\begin{flushleft}
			\textbf{Input}: $f$,\ $x$,\ $c$,\ $n$,\ $p$,\ $\gamma$ \\
			\textbf{Output}: $M$ \\
		\end{flushleft}
		\begin{algorithmic}[1] 
			\STATE Initialize $M_c^n(x),\ M,\ \mathbb{G}$
			\FOR{$i \leftarrow 1\enspace \textbf{to}\enspace n\ $}
			\STATE $M_c(x, \gamma_i) \leftarrow \nabla f_c(\gamma_i(x))$
			\STATE Add $\ M_c(x, \gamma_i) \enspace \text{to}\enspace M_c^n(x)$
			\CL{AbsoluteGrad }
			\STATE $M \leftarrow M + |M_c(x, \gamma_i)|$
			\ENDFOR
			\CL{Get Variance}
			\STATE $\mathbb{G} \leftarrow \mathbb{V}(M_c^n(x))$
			\CL{Filter the Variance as Gudie}
			\STATE \textit{p\_value} $\leftarrow$ percentile$(\mathbb{G},\ p)$
			\FOR{$\textbf{all}\enspace \textit{guide\_value} \in \mathbb{G}\ $}
			\IF{$\textit{guide\_value} \ge \textit{p\_value}$}
			\STATE $\textit{guide\_value} \leftarrow 1$
			\ENDIF
			\ENDFOR
			\STATE \textbf{return} normalize$(M \times \mathbb{G})$
		\end{algorithmic}
	\end{algorithm}
\end{minipage}%
\hspace{.3cm}
\begin{minipage}[t][][b]{0.5\linewidth}
	\begin{figure}[H]
		\centering
		\includegraphics[width=.9\linewidth]{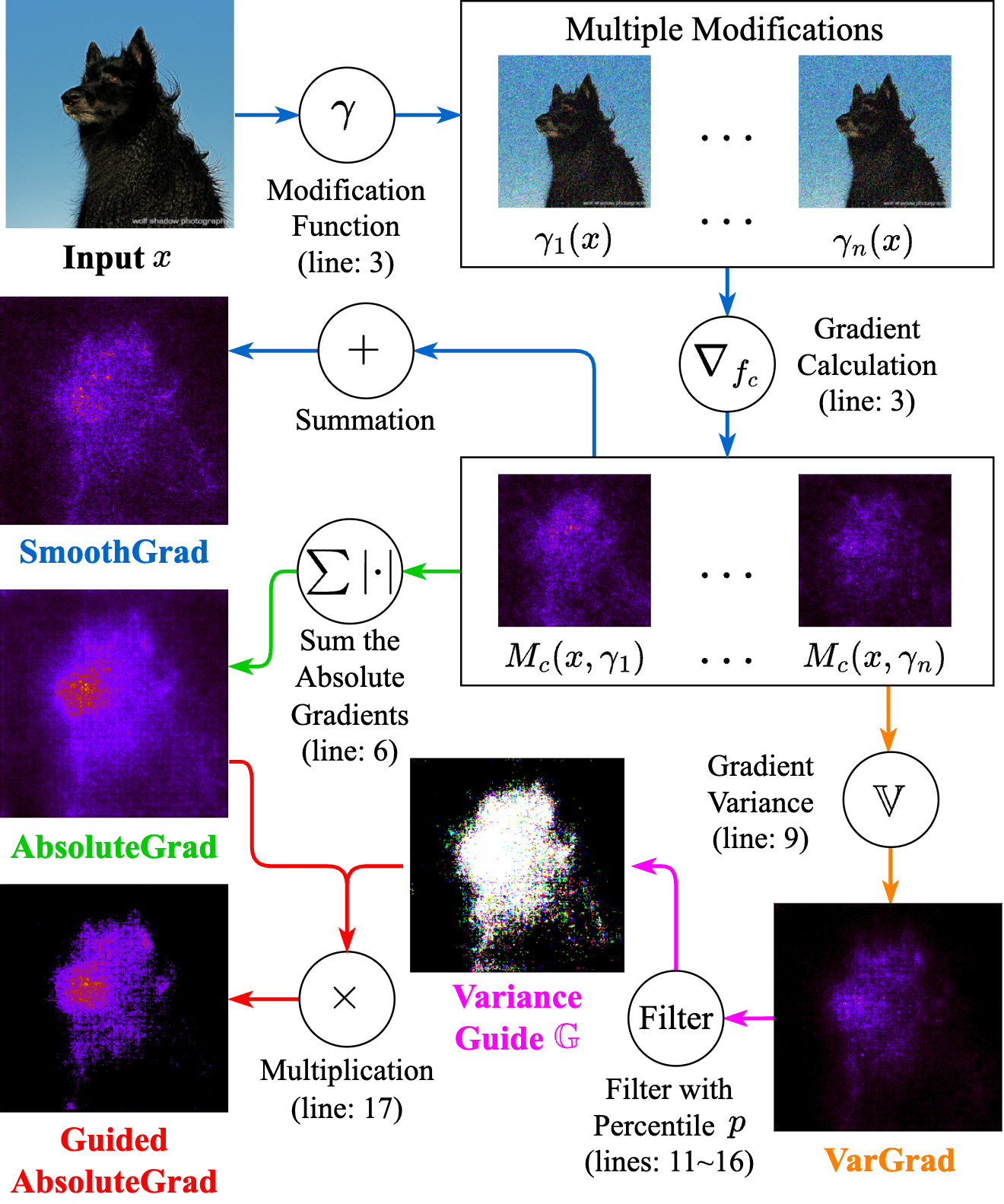}
		\caption{
			Illustrations for Guided AbsoluteGrad Algorithm.
		}
		\label{fig:gag_process}
	\end{figure}
\end{minipage}

\section{Define the ReCover And Predict}
In this section, we define the \textmc{ReCover And Predict (RCAP)} metric with notations and formulas.
We then discuss two propositions to describe the idea behind \textmc{RCAP}.

\subsection{Area Separation for Evaluation}
\label{sec:area_obj}
Metrics of \textmc{DAUC\&IAUC}~\cite{RISE} or \textmc{ROAR}~\cite{roar} evaluate all features' saliency, whereas the \textccpt{RCAP} could be more focused on the important area.
We first define the \textccpt{Focus Area} and \textccpt{Noise Area} of the saliency explanations.
Features with saliency values higher than the \textit{lower\_bound} percentile value in \textccpt{Focus Area}; otherwise, in \textccpt{Noise Area}.
We then define the \textccpt{Ground-truth Area} and \textccpt{Background Area}.
Features that only describe the target object are in the \textccpt{Ground-truth Area}; otherwise, in \textccpt{Background Area}.
For instance, the ground-truth segmentation annotated by humans is the perfect representation of the \textccpt{Ground-truth Area}.
We will define these areas in detail with notations in the next section.

\begin{figure}[ht]
	\centering
	\includegraphics[width=\linewidth]{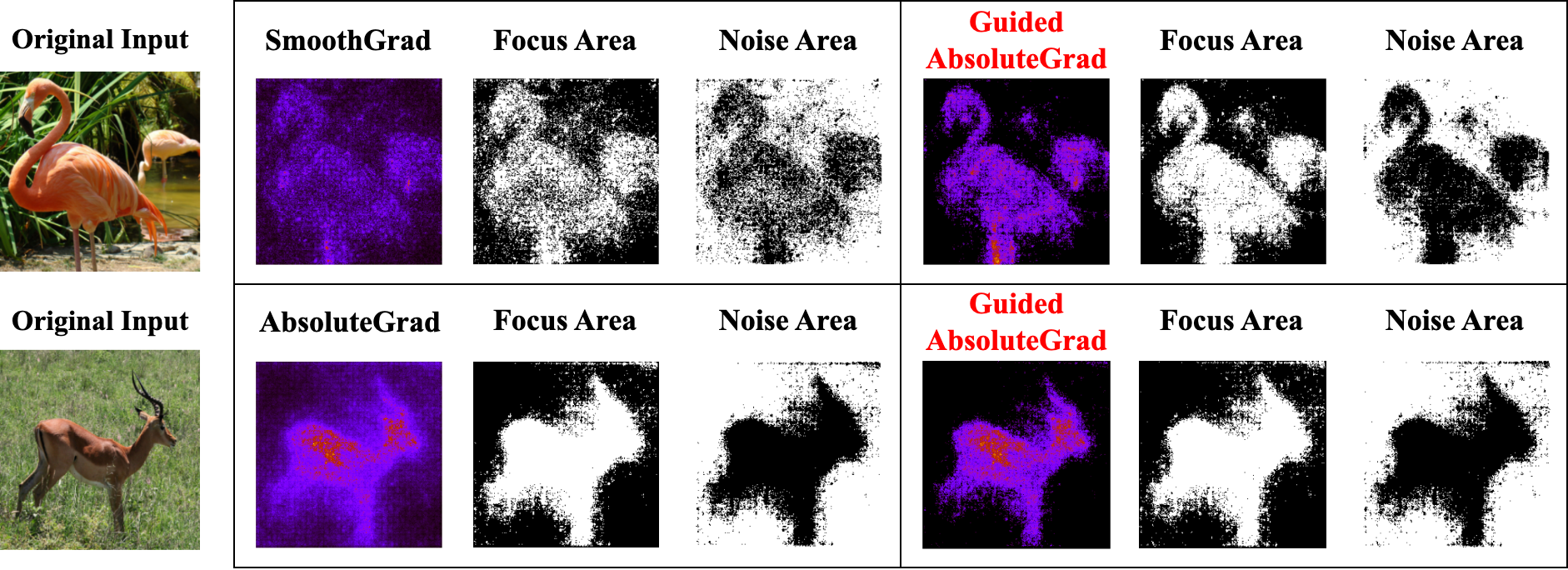}
	\caption{
		The \textccpt{Focus/Noise Area} separation for evaluating the gradient-based methods of SmoothGrad, AbsoluteGrad, and Guided AbsoluteGrad.
	}
	\label{fig:areas}
\end{figure}

Figure~\ref{fig:areas} shows the area separation of two images.
We use a \textit{lower\_bound} of 60$^{th}$ percentile to separate the areas.
The white pixels in the area indicate where we keep the saliency value.
The first row shows the \textccpt{Focus Area} of SmoothGrad is sparse, whereas the Guided AbsoluteGrad is dense.
We argue that the saliency map evaluation can be done in just the \textccpt{Focus Area}.
Both \textmc{DAUC\&IAUC} and \textmc{ROAR} evaluations are based on the prediction score of the gradually recovered samples.
Therefore, including more features from the \textccpt{Noise Area} will bring two results:
(1) prediction scores vary in a small range, indicating the most important features are explained already;
(2) prediction scores increased a lot, indicating the highest 40\percenting\ of saliency can not well cover the target object.
Both results are considered poor explanations and, thus, can be skipped.
We argue that evaluating the \textccpt{Noise Area} can be done without model prediction calls by just taking the saliency ratio of the \textccpt{Focus Area} with the entire saliency map.
The second row shows the saliency values in the \textccpt{Focus Area} of method AbsoluteGrad and Guided AbsoluteGrad are close to each other,
but the AbsoluteGrad assigns more saliency value in the \textccpt{Noise Area}.
It showcases that when the two saliency maps explain the target object equally well, we should consider the noise level.

\subsection{Metric Definition}

Based on the \textccpt{Ground-truth/Background Area} in Section~\ref{sec:gag_def} and the discussion of Figure~\ref*{fig:areas} in Section~\ref{sec:area_obj},
we define two objectives of the saliency map evaluation as follows.
\begin{itemize}
	\item \textccpt{Localization}:
	      It evaluates if the explained \textccpt{Focus Area} covers more on the \textccpt{Ground} \textccpt{-truth Area} and less on the \textccpt{Background Area}.
	\item \textccpt{Visual Noise Level}:
	      It evaluates if the saliency map can explain more saliency values in the \textccpt{Focus Area} and less in the \textccpt{Noise Area}.
\end{itemize}

Suppose the range of the saliency value is normalized to $[0,1]$, and $M$ is the saliency value matrix where $\sum M > 0$.
We take features from the highest salience value to \textit{lower\_bound} and partition them with the given interval \textit{interval}.
Then, starting from a baseline image (a black image, for instance), we recover the features multiple times with these partitions and obtain multiple recovered images.
With these partitions, we present the following notations:
\begin{itemize}
	\item $P = \{p_k\ |\ k \in \{1,2,\dots,j\} \}$ where $p_k$ is the $(100 - k\ *\ \textit{interval})^{th}$ percentile of the saliency values, \textit{interval} indicates a fixed difference of the adjacent percentile, and $k$ is the ordinal number of elements in $P$.
	      Note that $100 \ge k\ *\ \textit{interval} \ge \textit{lower\_bound} \ge 0$ where \textit{lower\_bound} is lowest percentile $p_k$ we take into consideration.
	      Then, we get the number of elements $j = (100 - \textit{lower\_bound})/\textit{interval}$.
	\item $I_{P} = \{I_{p_k}\ |\ k \in \{1,2,\dots,j\}\}$ where $I_{p_k}$ is the recovered image, which recovers the features that hold saliency value greater or equal to the $p_k^{th}$ percentile.
	\item $M_{P} = \{M_{p_k}\ |\ k \in \{1,2,\dots,j\}\}$ where $M_{p_k}$ is the subset of the saliency value of the recovered portion.
	      The last element $M_{p_j}$ indicates the \textccpt{Focus Area}.
	      Moreover, we use the hat notion $\hat{M}_{p_j} = (M - M_{p_j})$ to represent the \textccpt{Noise Area}.
	\item $\sigma(f_c(I_{p_k}))$ indicates softmax operation on prediction score of $I_{p_k}$ in class $c$.
	\item $\sum M_{p_k}/\sum M$ is the saliency ratio of the recovered part and the entire saliency map.
\end{itemize}

We then formulaize our \textmc{RCAP} metric as $RCAP(M)$ as:
\begin{equation}
	RCAP(M) = \frac{1}{j}\displaystyle\sum_{k=1}^j \bigg[\overbrace{\Big(\frac{\sum M_{p_k}}{\sum M}\Big)}^{\mathclap{\text{Visual Noise Level}}} \times \underbrace{\sigma(f_c(I_{p_k}))}_{\mathclap{\text{Localization}}} \bigg]
	\label{eq:eval}
\end{equation}

We interprate that the first term $\sum M_{p_k}/\sum M = \sum M_{p_k}/(\sum M_{p_k} + \sum \hat{M}_{p_k})$ measures the \textccpt{Visual Noise Level} with the saliency ratio of partition and the saliency map $M$.
If $\sum M_{p_k}$ gets larger, the ratio will get higher and approach 1, indicating less \textccpt{Visual Noise Level} of the explained \textccpt{Focus Area}.
The second term $\sigma(f_c(I_{p_k}))$ is the prediction confidence that measures the \textccpt{Localization}.
If more features from the \textccpt{Background Area} are recovered,
it means the explained saliency deviates from the target object, resulting in a smaller prediction score.

\subsection{Metric Proposition for Explanation Performance}
\label{sec:prop}
\begin{figure}[ht]
	\centering
	\includegraphics[width=\linewidth]{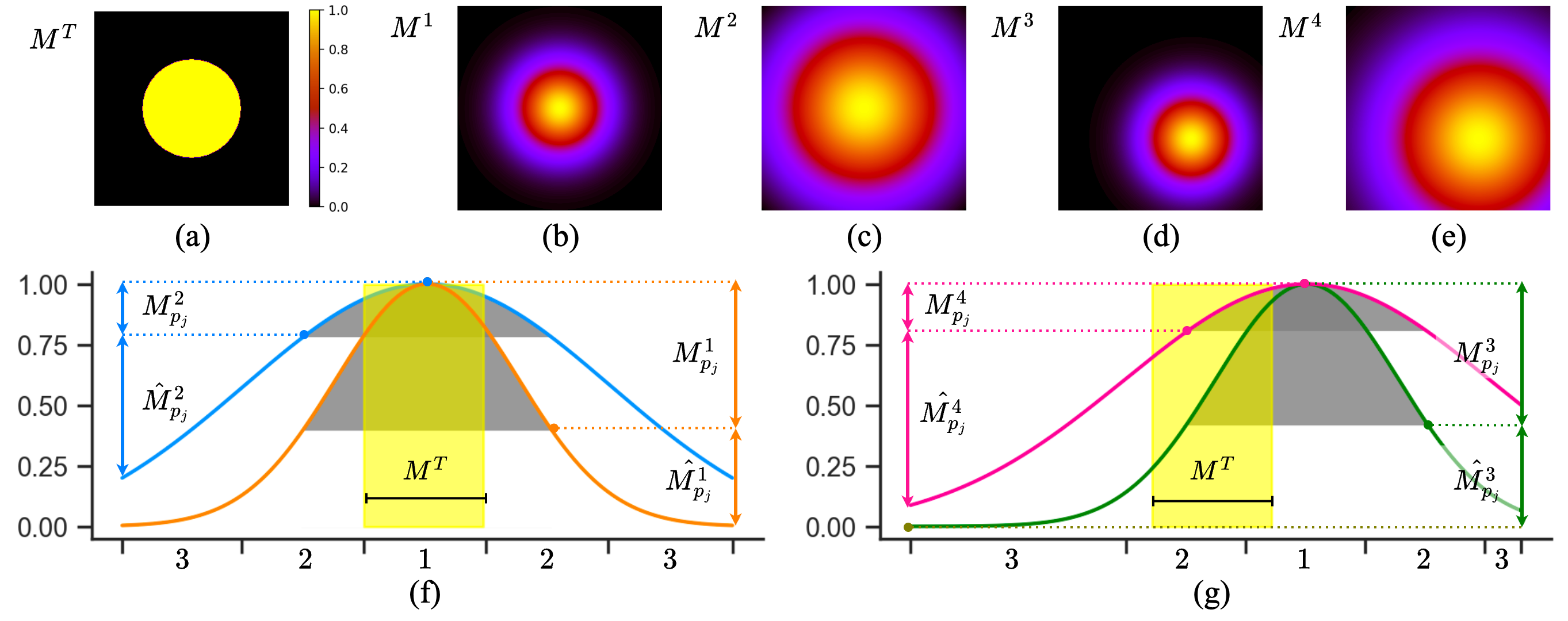}
	\caption{
		\textit{(a)} The ground-truth segmentation where the centered 20\percenting\ (yellow area) of the features is the target object.
		\textit{(b-e)} The four saliency maps with Gaussian distribution.
		\textit{(f)} The saliency distribution of map $M^1$ (orange line) and $M^2$ (blue line).
		\textit{(g)} The saliency distribution of map $M^3$ (green line) and $M^4$ (pink line).
		The gray area under each curve indicates the \textccpt{Focus Area} ($M_{p_j}$), and the yellow area is \textccpt{Ground-truth Area} ($M^T$).
		The $x$ axis is the partition number with $interval = 20\%$, resulting in five partitions.
		The $y$ axis is the saliency value ranged in $[0.0, 1.0]$.
	}
	\label{fig:coverage}
\end{figure}

We propose two propositions to describe how \textccpt{RCAP} evaluates the saliency map.
We denote the \textccpt{Ground-truth Area} as matrix $M^T$.
Additional high saliency value assigned outside $M^T$ is considered \textbf{a noise}.
We start by presenting the case in Figure~\ref{fig:coverage} to help understand the propositions better.
Suppose we use the $lower\_bound = 60^{th}$ for area separation,
indicating the highest 40\percenting\ of the saliency values are in the \textccpt{Focus Area} $M_{p_j}$.

Figure~\ref{fig:coverage} presents an ideal case where four saliency maps conform to Gaussian distribution.
$M^1$ is the best explanation with more precise \textccpt{Localization} and low \textccpt{Visual Noise Level}.
$M^2$ has the same distribution center as $M^1$ but with a higher standard deviation, resulting in higher noise levels.
$M^3$ has the same standard deviation, but its center deviates from the $M^T$ with a 30\percenting\ of the total width and height towards the lower right.
$M^4$ has the same standard deviation as $M^2$ and same center as $M^3$, resulting in the worst explanation.

We now define two propositions utilizing Figure~\ref{fig:coverage}.
Suppose we have two saliency maps $M^a$ and $M^b$ for class $c$ and evaluate them with the same \textit{interval} and \textit{lower\_bound}.

\begin{propbox}[label={prop:1}]{}
	For any $k \in \{1,2,\dots,j\}$,
	if $\sum M^a_{p_k} \approxeq \sum M^b_{p_k}$ and $\sum M^a \approxeq \sum M^b$,
	then we have:$\sum M^a_{p_k}/\sum M^a \approxeq \sum M^b_{p_k}/\sum M^b$.
	If the \textccpt{Focus Area} of $M^a$ covers more \textccpt{Ground-truth Area} than $M^b$ does,
	we should get $RCAP(M^a) > RCAP(M^b)$.
\end{propbox}

\textit{Proposition 1} describes how to evaluate the saliency map when they have similar \textccpt{Visual Noise Levels}.
This can be described by the $\sum M^1_{p_k}/\sum M^1 \approxeq \sum M^3_{p_k}/\sum M^3$ from Figure~\ref{fig:coverage}.
Now, the evaluation depends on the coverage $M_{p_j} \cap \ M^T$.
As can be observed, $M^1$'s \textccpt{Focus Area} (gray area) covers more \textccpt{Ground-truth Area} (yellow area) than $M^3$.
However, ground-truth explanation collection is not always available.
We hypothesize that the prediction confidence $\sigma(f_c(I_{p_k}))$ of the recovered partitions is approximate to this coverage.
We design experiments to validate this hypothesized measurement with the following steps.
Given the saliency map, we generate a reversed variant that swaps between the high and low saliency values.
Firstly, we retain the highest $l$\percenting\ of saliency values to ensure a certain level of explanation accuracy.
Secondly, we swap the saliency value in $[0, m^{th}]$ percentile with the saliency values in $[(100 - l - m)^{th}, (100 - l)^{th}]$ percentile.
With such modification, we make the \textccpt{Focus Area} of this saliency map cover more on the \textccpt{Background Area} and less on the \textccpt{Ground-truth Area}.
This results in a lower probability on $\sigma(f_c(I_{p_k}))$ than its original.

\begin{propbox}[label={prop:2}]{}
	For any $k \in \{1,2,\dots,j\}$, assuming $I^a_{p_k}$ recovers similar feature areas as $I^b_{p_k}$ does,
	then we have: $\sigma(f_c(I^a_{p_k})) \approxeq \sigma(f_c(I^b_{p_k}))$.
	If $M^b$ is visually noisier than $M^a$, we should get $RCAP(M^a) > RCAP(M^b)$.
\end{propbox}

\textit{Proposition 2} describes how to evaluate the saliency map explanations when map $M^a$ and map $M^b$ perform close on covering the \textccpt{Ground-truth Area}.
We leverage $M^1$ and $M^2$ from Figure~\ref{fig:coverage} to define the noise level.
Figure~\ref{fig:coverage} reveals that
(1) both maps have full coverage on $M^T$, resulting $\sigma(f_c(I_{p_k}^1)) \approxeq \sigma(f_c(I_{p_k}^2))$;
(2) the ratio of the area under the curve of $\sum M^1_{p_k}/(\sum M^1_{p_k} + \sum \hat{M^1}_{p_k})$ is greater than $\sum M^2_{p_k}/(\sum M^2_{p_k} + \sum \hat{M^2}_{p_k})$;
(3) the map $M^2$ has a higher saliency deviation, resulting in increased saliency in the \textccpt{Noise Area}.
Moreover, (2) and (3) are sufficient and necessary conditions for each other.
We therefore derive $(\sum M^1_{p_k} / \sum M^1) \times \sigma(f_c(I_{p_k}^1)) > \sum (M^2_{p_k} / \sum M^2) \times \sigma(f_c(I_{p_k}^2))$.
Hence $RCAP(M^1) > RCAP(M^2)$ is guaranteed based on Equation~\ref{eq:eval}. Thus, the \textit{Proposition 2} is proved.

Existing recovering/removing-based metrics such as \textmc{DAUC\&IAUC}~\cite{RISE} or \textccpt{ROAR}~\cite{roar} also hold the \textit{Proposition 1}.
However, they are \textbf{not sensitive} to the \textccpt{Visual Noise Level} change as is described in the \textit{Proposition 2}.
They only consider the rank of the partitional saliency map.
In the extreme case of Figure~\ref{fig:coverage}, $M^1$ and $M^2$ get the same evaluation from \textmc{DAUC\&IAUC},
regardless of the difference of saliency deviation,
as long as the same ranked saliency covers the same feature partitions.
We will showcase this limitation of \textmc{DAUC\&IAUC} in the experiment section.
We argue that \textit{Proposition 2} is also crucial to the saliency map evaluation.

\section{Experiment Setup}
\label{sec:exp}


\label{sec:exp_data_model}
We use \textmc{ReCover And Predict (RCAP)} with $\textit{lower\_bound} = 50^{th}$ and $\textit{interval} = 10$\percenting\ and the \textmc{DAUC\&IAUC}~\cite{RISE} for all the cases.
Additionally, we use metric {\textmc{Log-Cosh Dice Loss} ($L_{lc-dce}$)}~\cite{logcoshdiceloss} and \textmc{Mean Absolute Error (MAE)} for ImageNet-S case that has available ground-truth segmentation.
We select the samples that have over 90\% prediction confidence on the ground-truth class since poor performance on the prediction might lead to uninterpretable explanation~\cite{lapuschkin2019unmasking, anders2022finding}.
We conduct experiments on the following three cases.

\textbf{Case 1: ImageNet-S with ResNet-50.}
We select the ImageNet-S~\cite{imagenet-s} that provides the ground-truth segmentation for the benchmark dataset ImageNet~\cite{imagenet}.
A total of 3,976 images with $256\times 256$ resolution are used in the experiments.
We select the ResNet-50~\cite{resnet50} model pre-trained by Pytorch.

\textbf{Case 2: ISIC with EfficientNet.}
We select the combined dataset~\cite{isic2017, isic2018,isic2019, isic2020} of the International Skin Imaging Collaboration (ISIC) challenge for the years 2018 to 2020.
A total of 3,000 images with $448\times 448$ resolution are used in the experiments.
Accordingly, we select the pre-trained EfficientNet from~\cite{efficientnet}.

\textbf{Case 3: Places365 with DenseNet-161.}
We select the Plac\-es365~\cite{densenet161} dataset along with its pre-trained DenseNet-161 model.
A total of 5,432 images with $224\times 224$ resolution are used in the experiments.


\subsection{Ablation Study for Parameter $p$ of the Guided AbsoluteGrad}
We explore the performance of the Guided AbsoluteGrad methods with different percentile $p$ from 15$^{th}$ (GAG-15) to 85$^{th}$ (GAG-85).
The Guided AbsoluteGrad method with no variance guide ($p=0$) is the AbsoluteGrad (GAG-0).
From Table~\ref{tb:eval_rs_0}, we observe that when the $p$ is 85$^{th}$ (GAG-85),
the Guided AbsoluteGrad achieves the best performance on the \textccpt{RCAP} metric against GAG-0 by increasing
41.66\percenting\ (ImageNet-S), 51.38\percenting\ (ISIC), and 44.24\percenting\ (Places365).
For the \textccpt{MAE} metric, the performance trends are the same as the \textccpt{RCAP} metric.
In general, the more saliency we filter by the variance guide, the more \textccpt{RCAP} performance we gain.
In contrast, the performance of \textccpt{DAUC\&IAUC} remains stable, indicating that they are insensitive to the noise level change and only focus on the \textccpt{Localization} of the saliency map.
This also indicates that applying the variance guide won't lose much performance of \textccpt{Localization} but improves the \textccpt{Visual Noise Level} of the explanations.

\section{Experiment Results}
We compare the Guided AbsoluteGrad (GAG) with seven SOTA gradient-based methods along with the Grad-CAM~\cite{gradcam},
including SmoothGrad (SG)~\cite{smoothgrad}, VarGrad~\cite{vargrad},
Guided Backpropagation (GB)~\cite{guidedbp},
Guided Integrated Gradients (Guided IG)~\cite{guidedig},
Integrated Gradients (IG)~\cite{ig}, BlurIG~\cite{blurig},
Vanilla Gradient (VG)~\cite{vanillagrad}.
We use the same number of modifications for all methods as 20.
Based on the ablation studies of Table~\ref{tb:eval_rs_0},
we set $p=85^{th}$ for all methods that use our variance guides in all cases.

\begin{table}[!hb]
	\renewcommand{\arraystretch}{1}
	\small
	\begin{center}
		\setlength\tabcolsep{0.6pt}
		\begin{tabular}{M{1.6cm} | P{1.07cm} P{1.07cm}  P{1.07cm}  P{1.07cm}  P{1.07cm} | P{1.07cm}  P{1.07cm}  P{1.07cm} | P{1.07cm}  P{1.07cm}  P{1.07cm} }
			\Xhline{3\arrayrulewidth}
			       & \multicolumn{5}{c|}{\bfseries ImageNet-S} & \multicolumn{3}{c|}{\bfseries ISIC} & \multicolumn{3}{c}{\bfseries Places365}                                                                                                                                                                       \\
			       & $L_{lc-dce}^{\downarrow}$                 & $MAE^{\downarrow}$                  & RCAP$^{\uparrow}$                       & DAUC$^{\downarrow}$ & IAUC$^{\uparrow}$ & RCAP$^{\uparrow}$ & DAUC$^{\downarrow}$ & IAUC$^{\uparrow}$ & RCAP$^{\uparrow}$ & DAUC$^{\downarrow}$ & IAUC$^{\uparrow}$ \\
			\hline
			GAG-0  & {\rdnn 0.246226}                          & {\rdnn 0.509218}                    & {\rdnn 0.308988}                        & {\rdnn 0.107451}    & {\rdbf 0.684256}  & {\rdnn 0.377488}  & {\rdbf 0.450519}    & {\rdbf 0.754289}  & {\rdnn 0.108280}  & {\rdbf 0.181929}    & {\rdbf 0.514909}  \\
			GAG-15 & {\rdnn 0.242078}                          & {\rdnn 0.510331}                    & {\rdnn 0.308646}                        & {\rdnn 0.107423}    & {\rdnn 0.684055}  & {\rdnn 0.374189}  & {\rdnn 0.452034}    & {\rdnn 0.754036}  & {\rdnn 0.108537}  & {\rdnn 0.182319}    & {\rdnn 0.512090}  \\
			GAG-25 & {\rdnn 0.240169}                          & {\rdnn 0.505856}                    & {\rdnn 0.322326}                        & {\rdnn 0.107348}    & {\rdnn 0.683809}  & {\rdnn 0.391952}  & {\rdnn 0.455257}    & {\rdnn 0.754123}  & {\rdnn 0.113312}  & {\rdnn 0.183024}    & {\rdnn 0.509891}  \\
			GAG-35 & {\rdnn 0.239065}                          & {\rdnn 0.500605}                    & {\rdnn 0.337917}                        & {\rdnn 0.107125}    & {\rdnn 0.683852}  & {\rdnn 0.411819}  & {\rdnn 0.458050}    & {\rdnn 0.753785}  & {\rdnn 0.118414}  & {\rdnn 0.183841}    & {\rdnn 0.506977}  \\
			GAG-45 & {\rdbf 0.237792}                          & {\rdnn 0.494825}                    & {\rdnn 0.357661}                        & {\rdnn 0.106905}    & {\rdnn 0.683572}  & {\rdnn 0.437174}  & {\rdnn 0.461553}    & {\rdnn 0.753263}  & {\rdnn 0.126252}  & {\rdnn 0.183963}    & {\rdnn 0.504265}  \\
			GAG-55 & {\rdnn 0.237920}                          & {\rdnn 0.488934}                    & {\rdnn 0.381494}                        & {\rdnn 0.106929}    & {\rdnn 0.682545}  & {\rdnn 0.466530}  & {\rdnn 0.463375}    & {\rdnn 0.753502}  & {\rdnn 0.134830}  & {\rdnn 0.184283}    & {\rdnn 0.502009}  \\
			GAG-65 & {\rdnn 0.240749}                          & {\rdnn 0.482750}                    & {\rdnn 0.404181}                        & {\rdnn 0.106110}    & {\rdnn 0.682881}  & {\rdnn 0.500190}  & {\rdnn 0.464263}    & {\rdnn 0.753094}  & {\rdnn 0.144916}  & {\rdnn 0.184098}    & {\rdnn 0.501427}  \\
			GAG-75 & {\rdnn 0.247922}                          & {\rdnn 0.477539}                    & {\rdnn 0.422528}                        & {\rdnn 0.105346}    & {\rdnn 0.682472}  & {\rdnn 0.536391}  & {\rdnn 0.463632}    & {\rdnn 0.753085}  & {\rdnn 0.151659}  & {\rdnn 0.183435}    & {\rdnn 0.500861}  \\
			GAG-85 & {\rdnn 0.267944}                          & {\rdbf 0.474767}                    & {\rdbf 0.437721}                        & {\rdbf 0.104200}    & {\rdnn 0.681461}  & {\rdbf 0.571426}  & {\rdnn 0.462047}    & {\rdnn 0.752674}  & {\rdbf 0.156186}  & {\rdnn 0.183537}    & {\rdnn 0.501765}  \\
			\hline
			\Xhline{3\arrayrulewidth}
		\end{tabular}
	\end{center}
	\caption{
		Ablation Studies for Guided AbsoluteGrad with different $p$ to Equation~\ref{eq:norm_variance_processing}.
		The boldface number indicates the best in that experiment.
	}
	\label{tb:eval_rs_0}
\end{table}

\subsection{Gradient-based Method Comparison and Proposition Observation}

Part 2 of Table~\ref{tb:eval_rs_1} shows that our GAG method outperforms other gradient-based methods with regards to \textccpt{RCAP}, \textccpt{$L_{lc-dce}$}, and \textccpt{MAE} evaluations in all cases, without considering the Grad-CAM.
For \textccpt{IAUC}, the GAG method ranks first (ImageNet-S), third (ISIC), and second (Places365), respectively.
Whereas the \textccpt{DAUC} of GAG outperforms others for ISIC case, it ranks lower than the average in other cases.
When comparing with the Grad-CAM method,
our GAG method gets the closest $L_{lc-dce}$ evaluation and \textccpt{IAUC} against other gradient-based methods for case ImageNet-S.
Moreover, the GAG gets a better \textccpt{RCAP} in the case ImageNet-S and the closest \textccpt{RCAP} in the other two cases against other gradient-based methods.

As is discussed in Section~\ref{sec:prop}, we aim to observe the \textit{proposition 1}.
We alter methods SG, IG, Guided IG, BlurIG, and GB to their reversed variants with superscript $R$.
From part 2\&3 of Table~\ref{tb:eval_rs_1},
we observe that all reversed variants get significantly worse performance in all metrics against their original methods, which matches the description of \textit{proposition 1}.

\begin{table}[!ht]
	\renewcommand{\arraystretch}{1}
	\small
	\begin{center}
		\setlength\tabcolsep{0.6pt}
		\begin{tabular}{M{1.6cm} | P{1.07cm} P{1.07cm}  P{1.07cm}  P{1.07cm}  P{1.07cm} | P{1.07cm}  P{1.07cm}  P{1.07cm} | P{1.07cm}  P{1.07cm}  P{1.07cm} }
			\Xhline{3\arrayrulewidth}
			                & \multicolumn{5}{c|}{\bfseries ImageNet-S} & \multicolumn{3}{c|}{\bfseries ISIC} & \multicolumn{3}{c}{\bfseries Places365}                                                                                                                                                                       \\
			                & $L_{lc-dce}^{\downarrow}$                 & $MAE^{\downarrow}$                  & RCAP$^{\uparrow}$                       & DAUC$^{\downarrow}$ & IAUC$^{\uparrow}$ & RCAP$^{\uparrow}$ & DAUC$^{\downarrow}$ & IAUC$^{\uparrow}$ & RCAP$^{\uparrow}$ & DAUC$^{\downarrow}$ & IAUC$^{\uparrow}$ \\
			\hline
			Grad-CAM        & {\rdnn 0.134827}                          & {\rdnn 0.533307}                    & {\rdnn 0.405913}                        & {\rdnn 0.149310	}   & {\rdnn 0.690723 } & {\rdnn 0.641036}  & {\rdnn 0.617663}    & {\rdnn 0.789734}  & {\rdnn 0.270829}  & {\rdnn 0.293480}    & {\rdnn 0.595581}  \\
			\hline
			GAG             & {\rdbf 0.267944}                          & {\rdbf 0.474767}                    & {\rdbf 0.437721}                        & {\rdnn 0.104200	}   & {\rdbf 0.681461 } & {\rdbf 0.571426}  & {\rdbf 0.462047 }   & {\rdnn 0.752674 } & {\rdbf 0.156186}  & {\rdnn 0.183537}    & {\rdnn 0.501765}  \\
			VarGrad         & {\rdnn 0.343050}                          & {\rdnn 0.487181}                    & {\rdnn 0.380860}                        & {\rdnn 0.103676	}   & {\rdnn 0.681283 } & {\rdnn 0.507876}  & {\rdnn 0.463837 }   & {\rdnn 0.753282 } & {\rdnn 0.122973}  & {\rdnn 0.184234}    & {\rdnn 0.494715}  \\
			SG              & {\rdnn 0.330746}                          & {\rdnn 0.499001}                    & {\rdnn 0.106728}                        & {\rdnn 0.077568	}   & {\rdnn 0.604721 } & {\rdnn 0.364411}  & {\rdnn 0.490366 }   & {\rdnn 0.744140}  & {\rdnn 0.047968}  & {\rdnn 0.147710}    & {\rdnn 0.474288}  \\
			IG              & {\rdnn 0.338247}                          & {\rdnn 0.506464}                    & {\rdnn 0.044868}                        & {\rdnn 0.085472	}   & {\rdnn 0.512035 } & {\rdnn 0.344120}  & {\rdnn 0.504229 }   & {\rdnn 0.740013}  & {\rdnn 0.025042}  & {\rdnn 0.155752}    & {\rdnn 0.404904}  \\
			GB              & {\rdnn 0.382623}                          & {\rdnn 0.489085}                    & {\rdnn 0.172062}                        & {\rdnn 0.074304	}   & {\rdnn 0.650321 } & {\rdnn 0.323745}  & {\rdnn 0.499964 }   & {\rdnn 0.741209 } & {\rdnn 0.111551}  & {\rdnn 0.160278}    & {\rdbf 0.586063}  \\
			Guided IG       & {\rdnn 0.355837}                          & {\rdnn 0.490389}                    & {\rdnn 0.089705}                        & {\rdbf 0.069034	}   & {\rdnn 0.598204 } & {\rdnn 0.356746}  & {\rdnn 0.493596 }   & {\rdnn 0.747145}  & {\rdnn 0.030106}  & {\rdbf 0.139110}    & {\rdnn 0.443488}  \\
			BlurIG          & {\rdnn 0.373369}                          & {\rdnn 0.495475}                    & {\rdnn 0.085276}                        & {\rdnn 0.081145	}   & {\rdnn 0.556277 } & {\rdnn 0.409689}  & {\rdnn 0.479040 }   & {\rdbf 0.763821}  & {\rdnn 0.060184}  & {\rdnn 0.167668}    & {\rdnn 0.488642}  \\
			VG              & {\rdnn 0.345874}                          & {\rdnn 0.503979}                    & {\rdnn 0.033987}                        & {\rdnn 0.129176	}   & {\rdnn 0.451061 } & {\rdnn 0.323745}  & {\rdnn 0.499964 }   & {\rdnn 0.741209}  & {\rdnn 0.013201}  & {\rdnn 0.256877}    & {\rdnn 0.309799}  \\
			\hline
			SG$^{R}$        & {\rdnn 0.349800}                          & {\rdnn 0.506888}                    & {\rdnn 0.015605}                        & {\rdnn 0.098985	}   & {\rdnn 0.378892}  & {\rdnn 0.327550}  & {\rdnn 0.548077}    & {\rdnn 0.684696}  & {\rdnn 0.013012}  & {\rdnn 0.169543}    & {\rdnn 0.367622}  \\
			IG$^{R}$        & {\rdnn 0.352210}                          & {\rdnn 0.512288}                    & {\rdnn 0.006085}                        & {\rdnn 0.114989	}   & {\rdnn 0.305586}  & {\rdnn 0.344336}  & {\rdnn 0.560152}    & {\rdnn 0.690429}  & {\rdnn 0.010586}  & {\rdnn 0.166523}    & {\rdnn 0.346516}  \\
			GB$^{R}$        & {\rdnn 0.397432}                          & {\rdnn 0.493827}                    & {\rdnn 0.019291}                        & {\rdnn 0.143109	}   & {\rdnn 0.338570}  & {\rdnn 0.325672}  & {\rdnn 0.550051}    & {\rdnn 0.685277}  & {\rdnn 0.023322}  & {\rdnn 0.212114}    & {\rdnn 0.385824}  \\
			Guided IG$^{R}$ & {\rdnn 0.367862}                          & {\rdnn 0.494889}                    & {\rdnn 0.010105}                        & {\rdnn 0.096910	}   & {\rdnn 0.351039}  & {\rdnn 0.340619}  & {\rdnn 0.543531}    & {\rdnn 0.678459}  & {\rdnn 0.011942}  & {\rdnn 0.174522}    & {\rdnn 0.344207}  \\
			BlurIG$^{R}$    & {\rdnn 0.383503}                          & {\rdnn 0.499044}                    & {\rdnn 0.007828}                        & {\rdnn 0.111227	}   & {\rdnn 0.315826}  & {\rdnn 0.383930}  & {\rdnn 0.515028}    & {\rdnn 0.687931}  & {\rdnn 0.016926}  & {\rdnn 0.193670}    & {\rdnn 0.347256}  \\
			\Xhline{3\arrayrulewidth}
		\end{tabular}
	\end{center}
	\caption{
		Evaluation result for gradient-based method comparison and \textit{Proposition 1} observation.
		The $\uparrow$/$\downarrow$ indicates a higher/lower value is better.
	}
	\label{tb:eval_rs_1}
\end{table}

\subsection{Ablation Study on Different Gradient Processing}
We discuss how the different gradient processing approaches affect explanations in different contexts (dataset and model).
For methods SG, IG, Guided IG, BlurIG, and GB,
we make \textbf{five variants of gradient processing}:
($+$) use positive gradients only;
($-$) use negative gradients only;
($A$) use absolute gradients;
(${G}$) use our proposed gradient variance guide for noise deduction;
(${GA}$) use both absolute gradients and our variance guide.
After evaluating different variants, we divide them by their original method's evaluation to measure how much improvement/deterioration they make.
We then average the evaluations of all methods case by case for each metric.
Table~\ref{tb:eval_rs_3} shows the results where 1.0 indicates the performance is almost the same as its original methods.
The positive gradient generally enhances performance, while the negative gradient leads to minimal degradation.
On the other hand, utilizing the absolute gradient, applying the variance guide, or a combination of both tends to have more performance improvement in general.
It also shows that our \textccpt{RCAP} metric is more sensitive to the noise level changes against the \textccpt{DAUC\&IAUC} metric.
In the ISIC case, the \textccpt{IAUC} metric can barely recognize the improvement of gradient processing variants (4) and (5) indicated by \textccpt{DAUC} and our \textccpt{RCAP} metrics.

\begin{table}[!ht]
	\renewcommand{\arraystretch}{1}
	\small
	\begin{center}
		\setlength\tabcolsep{0.6pt}
		\begin{tabular}{M{1.3cm} | P{1.1cm} P{1.1cm}  P{1.1cm}  P{1.1cm}  P{1.1cm} | P{1.1cm}  P{1.1cm}  P{1.1cm} | P{1.1cm}  P{1.1cm}  P{1.1cm} }
			\Xhline{3\arrayrulewidth}
			     & \multicolumn{5}{c|}{\bfseries ImageNet-S} & \multicolumn{3}{c|}{\bfseries ISIC} & \multicolumn{3}{c}{\bfseries Places365}                                                                                                                                                                       \\
			     & $L_{lc-dce}^{\downarrow}$                 & $MAE^{\downarrow}$                  & RCAP$^{\uparrow}$                       & DAUC$^{\downarrow}$ & IAUC$^{\uparrow}$ & RCAP$^{\uparrow}$ & DAUC$^{\downarrow}$ & IAUC$^{\uparrow}$ & RCAP$^{\uparrow}$ & DAUC$^{\downarrow}$ & IAUC$^{\uparrow}$ \\
			\hline
			$+$  & {\rdtt 1.01}                              & {\rdtt 1.00}                        & {\rdtb 1.71}                            & {\rdtb 0.94}        & {\rdtb 1.05 }     & {\rdtt 1.00}      & {\rdtb 0.96}        & {\rdtt 1.00}      & {\rdtb 2.26}      & {\rdtb 0.89}        & {\rdtb 1.10}      \\
			$-$  & {\rdtt 1.01}                              & {\rdtt 1.00}                        & {\rdtb 1.34}                            & {\rdtt 1.08}        & {\rdtt 0.98 }     & {\rdtt 0.98}      & {\rdtt 1.02}        & {\rdtt 0.98}      & {\rdtb 1.73}      & {\rdtt 1.11}        & {\rdtt 0.98}      \\
			$A$  & {\rdtb 0.91}                              & {\rdtt 1.01}                        & {\rdtb 2.00}                            & {\rdtt 1.05}        & {\rdtb 1.11 }     & {\rdtt 0.98}      & {\rdtb 0.98}        & {\rdtb 1.01}      & {\rdtb 1.64}      & {\rdtt 1.03}        & {\rdtb 1.12}      \\
			$G$  & {\rdtt 1.06}                              & {\rdtb 0.98}                        & {\rdtb 2.62}                            & {\rdtb 0.96}        & {\rdtb 1.13 }     & {\rdtb 1.35}      & {\rdtb 0.99}        & {\rdtt 1.00}      & {\rdtb 3.20}      & {\rdtb 0.97}        & {\rdtb 1.20}      \\
			$GA$ & {\rdtb 0.97}                              & {\rdtb 0.97}                        & {\rdtb 2.86}                            & {\rdtb 0.99}        & {\rdtb 1.15 }     & {\rdtb 1.35}      & {\rdtb 0.98}        & {\rdtt 1.00}      & {\rdtb 3.32}      & {\rdtt 1.01}        & {\rdtb 1.21}      \\
			\hline
			\Xhline{3\arrayrulewidth}
		\end{tabular}
	\end{center}
	\caption{
		Average Performance Improvement/Degradation of different method variations of gradient processing.
		The boldface number indicates the result of improvement.
	}
	\label{tb:eval_rs_3}
\end{table}


\section{Conclusion}
In this paper, we propose a novel gradient-based XAI named Guided AbsoluteGrad.
The approach utilizes both positive and negative gradient magnitudes to aggregate saliency maps, employing gradient variance as a guide to reduce the noise of explanations.
We conduct ablation studies to reveal the performance influence of difference gradient processing approaches.
The result indicates that our assumption on the gradient magnitude matters to the saliency map explanation is valid.
Additionally, we propose a novel evaluation metric called \textmc{ReCover And Predict (RCAP)}.
We introduce two evaluation objectives: \textccpt{Localization} and \textccpt{Visual Noise Level}.
We further present how the \textccpt{RCAP} evaluates these objectives with proven propositions and empirical experiments.
We compare the Guided AbsoluteGrad method with seven SOTA gradient-based methods.
The experiment results show that the Guided AbsoluteGrad outperforms other SOTA gradient-based methods in the \textccpt{RCAP} metric and ranks in the front-tier in other metrics.



\printbibliography[heading=subbibintoc]

\end{document}